\documentclass[sigconf]{acmart}
\usepackage{hyperref}
\usepackage{url}
\usepackage{booktabs}       
\usepackage{amsfonts}       
\usepackage{nicefrac}       
\usepackage{microtype}      
\usepackage{xcolor}         
\usepackage{bm}
\usepackage{wrapfig}
\usepackage{graphicx}
\usepackage{amsmath}
\usepackage{caption}
\usepackage{grffile}
\usepackage{graphics}
\usepackage{amsthm}
\usepackage{thmtools}
\usepackage{thm-restate}
\usepackage{multirow}

\DeclareMathOperator*{\argmin}{arg\,min}
\AtBeginDocument{%
  }

\setcopyright{acmlicensed}
\copyrightyear{2018}
\acmYear{2018}
\acmDOI{XXXXXXX.XXXXXXX}
\acmConference[KDD '25]{Knowledge Discovery and Data Mining}{June 03--05,
  2018}{Woodstock, NY}
\acmISBN{978-1-4503-XXXX-X/2018/06}

\begin{document}

\title{Dynamic Mixture-of-Experts for Incremental Graph Learning}

\author{Lecheng Kong}
\email{jkong@amazon.com}
\affiliation{%
  \institution{Amazon}
  \country{}
}

\author{Theodore Vasiloudis}
\email{thvasilo@amazon.com}
\affiliation{%
  \institution{Amazon}
  \country{}
}
\author{Seongjun Yun}
\email{sjyun@amazon.com}
\affiliation{%
  \institution{Amazon}
  \country{}
}
\author{Han Xie}
\email{hanxie@amazon.com}
\affiliation{%
  \institution{Amazon}
  \country{}
}
\author{Xiang Song}
\email{xiangsx@amazon.com}
\affiliation{%
  \institution{Amazon}
  \country{}
}
\renewcommand{\shortauthors}{Kong et al.}

\begin{abstract}
  Graph incremental learning is a learning paradigm that aims to adapt trained models to continuously incremented graphs and data over time without the need for retraining on the full dataset. However, regular graph machine learning methods suffer from catastrophic forgetting when applied to incremental learning settings, where previously learned knowledge is overridden by new knowledge. Previous approaches have tried to address this by treating the previously trained model as an inseparable unit and using regularization, experience replay, and parameter isolation to maintain old behaviors while learning new knowledge.
These approaches, however, do not account for the fact that previously acquired knowledge at different timestamps contributes differently to learning new tasks. Some prior patterns can be transferred to help learn new data, while others may deviate from the new data distribution and be detrimental. Moreover, in the graph context, a node's receptive field contains neighbors from different data blocks, requiring variable processing, and an inseparable unit fails to account for such variability. To address this, we propose a dynamic mixture-of-experts (DyMoE) approach for incremental learning. Specifically, a DyMoE GNN layer adds new expert networks specialized in modeling the incoming data blocks. We design a customized regularization loss that utilizes data sequence information so existing experts can maintain their ability to solve old tasks while helping the new expert learn the new data effectively. As the number of data blocks grows over time, the computational cost of the full mixture-of-experts (MoE) model increases. To address this, we introduce a sparse MoE approach, where only the top-$k$ most relevant experts make predictions, significantly reducing the computation time. Our model achieved 4.92\% relative accuracy increase compared to the best baselines on class incremental learning, showing the model's exceptional power.
\end{abstract}



\received{20 February 2007}
\received[revised]{12 March 2009}
\received[accepted]{5 June 2009}

\maketitle

\section{Introduction}
Graph neural networks (GNN) achieved great success in modeling graph data and have many applications, such as recommender systems~\citep{recommender}, drug discovery~\citep{drug}, and traffic forecasting~\citep{traffic}. However, in many real-world settings, the graph is dynamic, starting small and expanding over time, and the training data arrive as sequences of data blocks with timestamps. Naive approaches train on the \textbf{full graph} whenever new data appears, which incurs expensive computational costs due to repetitive training on old data. On the other hand, simply finetuning conventional GNNs on the new data leads to catastrophic forgetting, where the model's prediction shifts toward the new data distribution and forgets how to handle previously learned tasks upon encountering new data~\citep{hpn, zhang2024topology, cui2023lifelong, graphsail}. This motivated a series of \textit{continual learning} research to tackle this problem~\citep{contsurvey, lifelongsurvey, distributionshiftsurvey}.

Pioneering efforts focused on adapting incremental learning approaches for other data modalities to the graph domain~\citep{er,graphsail,riegrace}. However, they ignore the fact that nodes and edges are not independent and identically distributed (i.i.d.) in the graph learning scenario~\citep{wang2022lifelong, wang2020streaming}. In the vision and language domain, individual image or text data points do not affect each other, and future data blocks do not impact the data distribution of the existing data blocks. In contrast, new graph data blocks connect to existing data via edges and could significantly change existing data distribution. For example, an incoming data block can add edges between two disconnected components in an existing graph, drastically changing the graph topology and, subsequently, the learned model behavior. Such uniqueness makes graph incremental learning an even more challenging scenario than incremental learning in other domains.

Subsequent efforts tackled the problem in several ways~\citep{tan2022graph, graphsail, wang2022lifelong}. For instance, PI-GNN~\citep{pi-gnn} rectified the old model on the graph modified by the new data. TWP~\citep{twp} identified topology-aware parameters to stabilize the model under graph structure shift. DiCGR~\citep{kou2020disentangle} breaks relation triplets to components to better capture graph structures. RLC-CN~\citep{rcl-cn} determines the optimal memory size for effective efficiency/performance trade-off. SSRM~\citep{ssrm} minimizes structural shift loss to mitigate performance degradation on old nodes.

These methods show improvements in the graph setting compared to the naive adaptation of incremental learning methods from other domains. However, a commonality of these approaches is that they build the new model upon an inseparable old model. Specifically, Elastic Weight Consolidation (EWC)~\citep{ewc} used the old model parameters as the single regularization target for all parameters; Experience Replay (ER)~\citep{er} trained the model using all saved subsets of nodes from old data blocks; Parameter Isolation (PI-GNN)~\citep{pi-gnn} froze all old model parameters and used an additional network to modify the model output. Such a pattern causes inflexibility when dealing with the "stability versus plasticity dilemma"~\cite{svpdilemma} commonly seen in the continual learning domain, where the model needs to effectively trade-off between maintaining old knowledge (stability) and learning new (plasticity).
\begin{figure*}
    \centering
    \includegraphics[width=\linewidth, trim={0.1cm, 8.8cm, 4.7cm, 0.5cm}, clip]{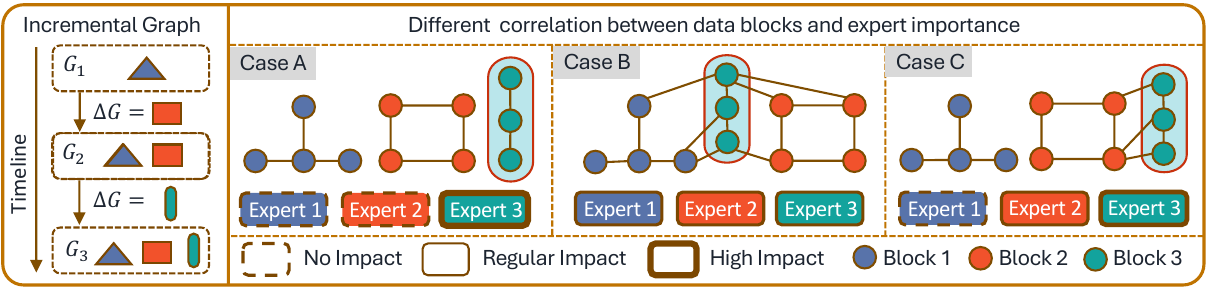}
    \caption{Left: Data blocks arrive in sequence. Right: Different connection types of three data blocks. Our proposed method activates dedicated experts when inferring relevant data blocks.}
    \label{fig:init_exp}
\end{figure*}

For example, in Figure~\ref{fig:init_exp}, blue, red, and green nodes represent data blocks one, two, and three that arrive in order in all three cases, and we update the model whenever a data block arrives. Block one and two are identical in all cases, while block three is isolated, connected to both blocks two and three, and connected to block two only in cases A, B, and C, respectively. After learning blocks one and two, the model needs to learn block three while maintaining old knowledge. Previous approaches tackle cases A and B, where the new block relates to old blocks by similar patterns (no correlation to all blocks in case A, and high correlation in case B), and they use the same strategy to retain knowledge from both blocks one and two. However, in case C, where the new block is partially dependent on the old blocks, they still apply the same stabilizing strength to knowledge acquired from previous blocks. While a stronger stabilizing effect is required to maintain knowledge in block one as it is more divergent from the new block, a smaller effect is desirable for knowledge in data block two as it is similar to the new block and weaker stabilizing enables better knowledge transfer and learning. Conventional approaches lack the flexibility to handle such diversity among data blocks and hence witness performance degradations.

To tackle this problem, we propose a Dynamic Mixture-of-Expert (DyMoE) module to use separate expert networks to model different data blocks with a gating mechanism to synthesize information from the most relevant experts. Specifically, the module has the same number of experts as trained data blocks, and the experts are dedicated to learning from their corresponding data blocks. DyMoE routes the input to experts through a gating mechanism, and the expert outputs are combined through the gating values by a weighted sum. This approach explicitly considers the correlation between different experts and data blocks. For the same example in Figure~\ref{fig:init_exp}, we train three separate experts with specialization in their corresponding data blocks. We then compute the relevance of the experts to the input. The experts with higher relevance have a higher impact on the prediction. This approach dynamically adjusts the combination of knowledge from different data blocks; less impactful experts are disabled during inference to reduce misleading information. When a new data block arrives, we append a new expert dedicated to the new data block without interfering with the knowledge of existing experts during training. To ensure each expert focuses on the assigned data block, we propose a block-guided loss as a training objective that enforces a high relevance score of experts to the input from their corresponding data blocks, greatly reducing catastrophic forgetting while allowing flexible querying of old knowledge.

While DyMoE applies to most neural architectures, \textbf{it is particularly useful for graph incremental learning}, where nodes' receptive fields contains neighbor nodes from different data blocks. We apply DyMoE to graph neural networks such that different experts can process nodes from different data blocks in the same ego subgraph, largely preserving the authenticity of the node's representation. Another challenge is that nodes in future data blocks change the topology and information in the old data blocks, as they become neighbor nodes to the old data blocks. We extend the gating mechanism to distinguish future neighbors from the old, we then filter future nodes for the old experts, largely recovering the old behavior and reducing forgetting. To increase efficiency, we propose a sparse variant, inspired by \citet{shazeer2017outrageously}, that only considers the most relevant experts during inference, significantly reducing the computation complexity while maintaining high accuracy. In this paper, we

\begin{itemize}
    \item Identified the issue of existing continual learning methods that ignore the correlation between different data blocks.
    \item Designed a DyMoE module with specialized experts for each data block and proposed data block-guided loss to minimize the negative interference between experts.
    \item Interleave the DyMoE module into GNNs and use graph block-guided loss to address the data shift problem unique to graph continual learning.
    \item Developed a sparse version of the DyMoE module so the model is both efficient and effective.
\end{itemize}

In our empirical evaluation, DyMoE maintains the same efficiency, while significantly improving over the best baseline in class incremental learning setting. The model also demonstrates strong results in instance incremental settings. 
\section{Preliminaries}
\textbf{Graph Incremental Learning.} This paper focuses on incremental learning for node classification. Specifically, we follow the widely adopted problem formulation~\citep{contsurvey, lifelongsurvey}, and aim to incrementally learn from a graph data block sequence $D=\{G^{(1)},...,G^{(t)}\}$, and each data block is a graph $G^{(i)}=(V^{(i)}, E^{(i)}, \bm{y}^{(i)})$ where $V^{(i)}$ is the set of nodes, and $E^{(i)}$ is the set of edges, and $\bm{y}^{(i)}$ is the classification labels of the nodes. Future graph snapshots expand on existing graphs, and $G^{(i)}$ is a subgraph of $G^{(j)}$ for $i<j$. We additionally use $\Delta G^{(i)} = (V^{(i)}\setminus V^{(i-1)}, E^{(i)}\setminus E^{(i-1)})$ to represent the graph delta between $G^{(i)}$ and $G^{(i-1)}$. We use $b(v)$ to indicate the data block index where the data/node $v$ first appears. In the incremental learning setting, data arrive in order, and the $i$-th model is only trained and evaluated on $(G^{(1)},...,G^{(i)})$ without any knowledge about future graphs. The goal is to maximize the overall accuracy on each data block while minimizing the performance drop on previous data blocks. If the classes in $\bm{y}^{(i)}$ persist throughout all blocks, we refer to the task as instance-incremental learning~\citep{van2022three}. If the classes in $\bm{y}^{(i)}$ are disjoint, we refer to the task as class-incremental. In this case, new data blocks also bring in new classes~\citep{zhang2022cglb}, and the model needs to classify a sample without knowing its corresponding block during inference.

The naive solution is to train a model on the full graph $G^{(i)}$ for every block. However, this requires retraining on all old data multiple times, incurring huge computational costs. Incremental learning methods aim to train only on the graph delta while maintaining good performance on the old data. 

To evaluate a model, let $a_{i,j}$ be the accuracy of all evaluation nodes in $G_i$, evaluated by the model after training $G_j$, which is a superset of evaluation nodes in $G^{(i)}$ and $i\leq j$. We evaluate the overall model performance by Average Accuracy (AA) and Average Forgetting (AF),
\begin{equation}
    AA=\frac{1}{t}\sum_{i=1}^t a_{i,i},\quad AF=\frac{1}{t}\sum_{j=1}^t \frac{1}{j}\sum_{i=1}^j a_{i,j}-a_{i,i}
\end{equation}
where $t$ is the number of data blocks. AA evaluates the model's average accuracy right after the model is trained on a data block, while AF evaluates the model's ability to retain knowledge from previous data blocks. The goal of an incremental learning method is to maximize AA and minimize AF.

\textbf{Graph Neural Networks}
Graph neural networks iteratively update a node's embeddings from their neighbor nodes through message-passing layers~\citep{gilmer2017neural}. Specifically, for a graph $G=(V, E)$, the $i$-th layer of a $T$-layer GNN is,
\begin{equation}
    \bm{h}_v^{(i+1)} = COMB(\bm{h}_v^{(i)}, AGGR(\{\bm{h}_u^{(i)}|u\in\mathcal{N}(v)\}), \quad v\in V
\end{equation}
where $\mathcal{N}(v)$ are the direct neighbors of $v$. Different GNN designs differ mainly by the combine (COMB) and aggregate (AGGR) functions.
\section{Dynamic Mixture-of-Experts Graph Neural Network}
\begin{figure*}
    \centering
    \includegraphics[width=\linewidth, trim={2cm 1cm 9cm 3cm}, clip]{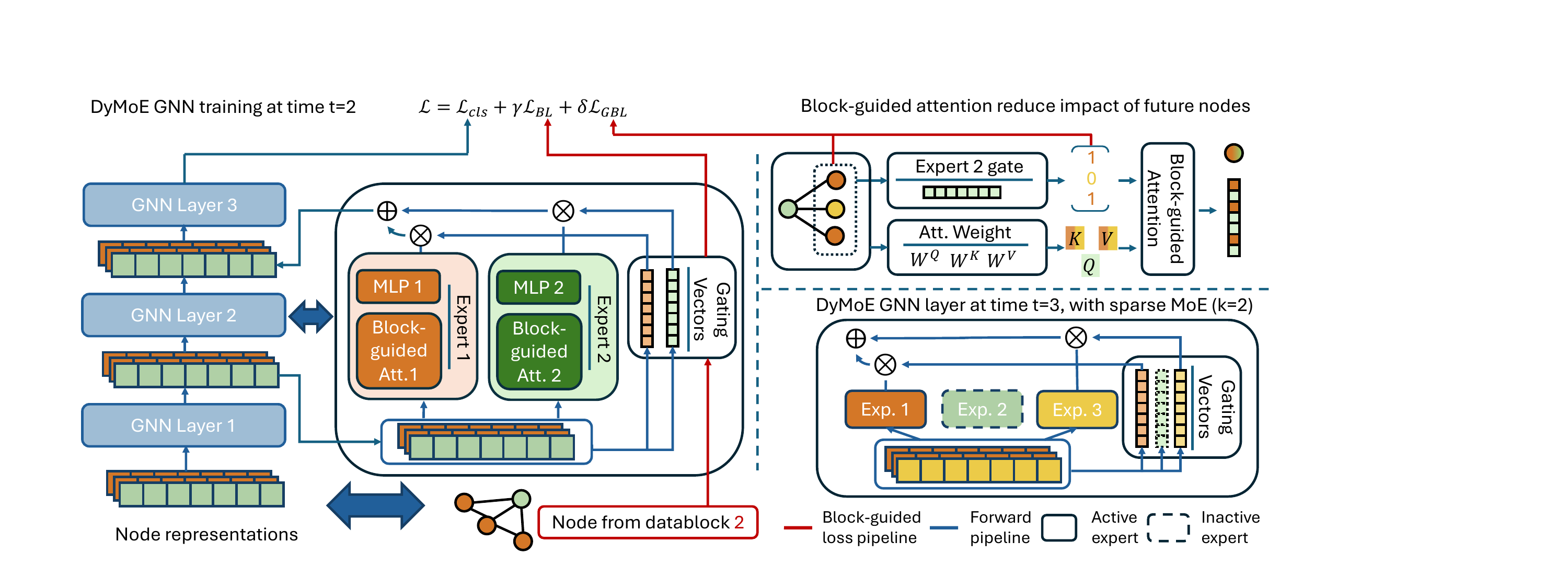}
    \caption{Pipeline of DyMoE GNN. Left: Each GNN layer has $t$ experts with individual attention and MLP weights. We compute gating values from the node representations and the gating vectors. During training, we compute a block-guided loss between the gating values and the data block index for correct expert selection. Top-Right: the graph block-guided loss assigns additional weights to neighbor nodes and filters unrecognized nodes for experts. Bottom-Right: When a new data block arrives, we add a new expert and a gating vector to the DyMoE module. In the sparse case, only the most important experts are used.}
    \label{fig:pipeline}
\end{figure*}
This section first introduces the Dynamic Mixture-of-Experts (DyMoE) module that dynamically increases the number of experts for new data blocks. We then describe the integration of DyMoE and GNN for effective graph incremental learning. To overcome the efficiency issue with long data sequences, we propose Sparse DyMoE to reduce the complexity of our framework. The overall architecture of the framework is shown in Figure~\ref{fig:pipeline}.

\subsection{Dynamic Mixture-of-Experts Module}\label{sec:dymoe}
Conventional mixture-of-experts (MoE) models create networks of the same architecture and apply a gating mechanism to combine the networks' outputs using a weighted sum~\citep{shazeer2017outrageously}. The number of experts is fixed after initialization. However, to accommodate new data blocks, MoE models suffer from the same issue as in other continual learning methods. They still need to adjust the weights of all previous experts, leading to forgetting. To mitigate this, we propose the DyMoE module, adding one expert for every new data block without modifying previously trained experts. Let $\mathcal{F}$ be a class of neural networks with the same architecture, and $f_\theta\in\mathcal{F}$ be an instance of the network parametrized by $\theta$. Specifically,
\begin{equation}
    \bm{h} = f_\theta(\bm{x})\quad \bm{x}\in\mathcal{R}^n, \bm{h}\in\mathcal{R}^m, f_\theta\in\mathcal{F}
\end{equation}
where $\bm{x}$ and $\bm{h}$ are the input and output to the network, and $n$ and $m$ are the input and output dimensions. Given an incremental data sequence $D=\{(X^{(1)}, \bm{y}^{(1)}),...,(X^{(t)}, \bm{y}^{(t)})\}$, DyMoE handles the first data block like a conventional network. Specifically, it minimizes the empirical loss,
\begin{equation}
    \argmin_{\theta_1} \frac{1}{|X^{(1)}|}\sum_{i,\bm{x}\in X^{(1)}} \mathcal{L}(\bm{y}^{(1)}_i, f_{\theta_1}(\bm{x}))
\end{equation}
The loss function $\mathcal{L}$ is task dependent, and we use cross-entropy loss for classification. For the second data block, we will add one expert and gating vectors to the overall model. To compute the output, we have 
\begin{equation}
\begin{split}
    \bm{h}&= f_{\{\theta_1,\theta_2\}}(x) = \alpha_1 f_{\theta_1}(x)+\alpha_2 f_{\theta_2}(x), \\
    \alpha_i &= \frac{exp(s(\bm{x}, \bm{g}_i))}{exp(s(\bm{x}, \bm{g}_1)) + exp(s(\bm{x}, \bm{g}_2))} \quad i\in\{1,2\}
\end{split}
\end{equation}

where $\bm{g}$ are gating vectors associated with each expert, $s(\cdot,\cdot)$ is a similarity measure, and we use softmax on the similarities to compute the importance of each expert for the input. Note that this formulation is the same as existing MoE approaches, and the key difference is that the number of experts dynamically increases as more data arrive. Subsequent data blocks follow the same procedure, where the output is computed as,
\begin{equation}\label{eq:moe}
    \bm{h}= f_{\{\theta_1,...,\theta_t\}}(\bm{x}) = \sum_{i=1}^t\alpha_i f_{\theta_i}(\bm{x}), \quad \alpha_i = \frac{exp(s(\bm{x}, \bm{g}_i))}{\sum_{j=1}^t exp(s(\bm{x}, \bm{g}_j))}
\end{equation}

When training on a new data block $t$, we only optimize the new expert and all the gating vectors, specifically, 
\begin{equation}
    \argmin_{\theta_t, \{\bm{g}_1...\bm{g}_t\}} \mathcal{L}_{cls},  \mathcal{L}_{cls}= \frac{1}{|X|}\sum_{i,\bm{x}\in X} \mathcal{L} (\bm{y}_i, f_{\{\theta_1,...,\theta_t\}}(\bm{x}))
\end{equation}
Intuitively, this training scheme completely preserves the knowledge obtained from previous data blocks. Ideally, when the gating vectors are perfectly trained to distinguish which data block a particular data point belongs to, the model can \textbf{fully recover} the output of that data point, eliminating forgetting.

\textbf{Block-guided loss.} While the experts can preserve learned knowledge, the new experts are randomly initialized and start with trivial predictions on all data. The model will rely on the existing trained experts to make predictions, though they may carry old, potentially suboptimal, knowledge regarding the new data block. The gating vectors, including the new one, will tend to select the old experts during training. The model can hence be trapped at the local minimum without properly training the new dedicated experts. Consequently, we need to inject the information about the correct experts for our dynamically initialized new modules. This is difficult in conventional MoE because of the lack of supervision for correct experts. However, in continual learning, data arrive in blocks, and since experts are designed to handle individual data blocks, \textit{we know exactly which expert a particular training data point should be assigned to}.
We propose a \textbf{block-guided loss} to train the gating vectors for correct expert assignment. Specifically, for an arbitrary data point $\bm{x}$, in addition to its classification loss, we add a cross-entropy loss between the gating values of all experts and the data point's corresponding data block index $b(\bm{x})$. The computation is valid because the number of experts equals the number of witnessed data blocks. The loss forces an expert's corresponding data and gating vector to have large similarities, maximizing the likelihood of using the correct expert to generate output for the data. Specifically,
\begin{equation}
    \mathcal{L}_{BL}(x)=CE(Softmax(\alpha_1...\alpha_t), OneHot(b(\bm{x}),t)), \bm{x}\in X
\end{equation}
where $CE$ is cross-entropy loss, $\alpha$ is the gating values, $OneHot(j,t)$ generates a $t$-dimensional one-hot vector whose $j$-th entry is one. Note that if we naively take $X$ as the new samples in the most recent data block $X^{(t)}$, all of them will have the same data block index (the last index), causing the model to always use the last expert. Hence, we store a small sample set from each data block as the memory set, $M^{(i)}\subset X^{(i)}$  and $|M^{(i)}|\ll |X^{(i)}|$, and take $X=\bigcup_{i=1}^{t-1} M^{(i)}\cup X^{(t)}$, so the model can adjust the gating values accordingly. The size of memory set $|M|=p|X|$ where $p<1$. The memory set construction details can be found in Appendix~\ref{app:mem}.

Note that we only use such information during training, and the model does not need the time information, or which block that a data point belongs to, during inference, making the model perfectly viable for difficult tasks such as class-incremental learning. The overall training loss is,
\begin{equation}
    \mathcal{L} = \mathcal{L}_{cls} + \gamma \mathcal{L}_{BL}
\end{equation}
where $\beta$ is a hyperparameter controlling the strength of regularization. The combined framework essentially attempts to train a data-block-dedicated classifier and out-of-distribution detectors for every data block. The gating mechanism gives high weight to in-distribution experts while minimizing the impact of out-distribution experts. While this approach applies to arbitrary data modality, it is particularly critical in the graph learning setting, where a node's neighbor might be from different data blocks and require different processing. We elaborate more on this in Section~\ref{sec:dygnn}. We theoretically show the advantages of our proposed model over the Parameter Isolation (PI)~\citep{pi-gnn} approach, a representative architectural approach for continual learning.
\begin{restatable}[]{thm}{mainthm}
\label{thm:main_thm}
For an arbitrary continual learning problem, suppose a PI model obtains a cross-entropy classification loss $\mathcal{L}_{PI}$, there exists a parametrization of DyMoE that achieves cross-entropy classification loss $\mathcal{L}_{Dy}= \mathcal{L}_{PI}$. When the data sequence follows a mixture of Gaussian distribution, we have $\mathcal{L}_{Dy}\leq \mathcal{L}_{PI}$.
\end{restatable}
The proof is in Appendix~\ref{sec:app_proof}. In the proof, we first show that DyMoE is at least as powerful as PI. We then show under the Gaussian Mixture assumption of the input data block sequence; the DyMoE obtains strictly lower loss, which shows the model's superiority.

In practice, the memory set is very small to ensure efficiency, but we jointly train on it with the full dataset from the new data block, which can give the model a biased understanding of the data distribution (i.e. most of the data are from the last data block). Hence, we propose a \textit{data balancing} training procedure, where, after the regular training epochs, we collect the memory set for the new data block, combine it with all previous training memory sets, and train a few epochs on them to reflect the actual distribution of the entire input sequence. We elaborate on the data balancing training procedure in Appendix~\ref{app:mem}.

\subsection{Dynamic Mixture-of-Expert Graph Neural Network}\label{sec:dygnn}
\begin{figure}[t]
    \centering
    \includegraphics[width=0.99\linewidth, trim={0cm 8.7cm 17.2cm 0.0cm},clip]{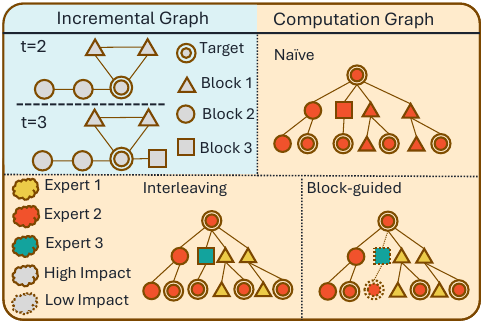}
    \caption{The computation graphs of the same target node at $t=3$ by different approaches.}
    \label{fig:gnnsep_exp}
    \vspace{-15pt}
\end{figure}
We then introduce fusing the DyMoE with a graph neural network. Note that the DyMoE module does not assume any specific network architectures, and a naive solution can treat a multi-layer GNN as $\mathcal{F}$. However, this ignores the unique property of graph data in continual learning, where a target node's neighbors are from different data blocks. For example, in Figure~\ref{fig:gnnsep_exp}, the target node is from data block two but is later connected to nodes in block three. The naive approach will assign expert two to process the target node and all of its neighbors, but its neighbor nodes are from blocks one and three, and expert two lacks the knowledge to properly handle them, leading to compromised performance. Ideally, a DyMoE model should assign neighbors to their corresponding experts. Hence, we propose interleaving the DyMoE modules into each GNN layer to correct this. We modify a transformer Graph Convolution Layer~\citep{transgnn} to the architecture of each expert layer.
Specifically,
\begin{equation}
     f_{\theta_t}^{(i)}=\bm{h}_{v,t}^{i}=MLP(\bm{h}_v^{(i-1)}+Att(\bm{h}_v^{(i-1)},\{\bm{h}_u^{(i-1)}|u\in \mathcal{N}(v)\}))\\
\end{equation}
where $Att$ is the attention mechanism with target node's feature as query and neighbor nodes' features as key and values, formally
\begin{equation}
\begin{split}
    &Att(\bm{h}, U)=softmax(\frac{\bm{q}K}{\sqrt{d}})V\\
    \bm{q}=W^Q_t\bm{h},K=&W_t^KU,V=W_t^VU, \quad W_t^Q,W_t^K,W_t^V\in \mathcal{R}^{n\times n}
\end{split}
\end{equation}
where $W$s are the attention weights, and each expert has individual MLP and attention weights. The interleaving DyMoE design can be easily extended to other GNN architectures. The key difference between this and naive approaches is that DyMoE naturally handles the problem where neighbor nodes of the target node are from different data blocks. As long as we properly train the gating vectors, DyMoE can route each neighbor node to their corresponding expert instead of the one corresponding to the target node. This approach generates the most authentic representations.

Another critical trait of incremental graph learning is that new data can change the existing graph's overall topology. In the same example, expert two is trained to handle data block two without data from block three. However, after data block three arrives, its data changes the neighborhood of the target node, meaning that even we correctly assign the experts, we will not recover the performance and output. Essentially, the modified topology in the graph causes the old model to shift from its original prediction and cause a performance decrease. 

To overcome this issue, we observe that since the gating mechanism introduced in Section~\ref{sec:dymoe} can be used to distinguish which data block a node is from, we can extend it to predict whether a node occurs before an expert is added to the model. Correctly predicting this target allows us to suppress future nodes' influence on old experts and thus maximally preserve the old behavior. Specifically, we create an additional gating vector $\bm{p}_i$ for each expert and compute gating values $\beta_{u,t}$ for node $u$ on expert $t$ as:
\begin{equation}
    \beta_{u,t} = sigmoid(\bm{p}_t\cdot (W^Ph_u)),\quad u\in G
\end{equation}
where $W^P$ is a linear projection shared by all experts in the same layer and $\beta$ is independent of other experts, and it indicates whether the node $u$ is incremented to the graph before expert $t$ is added to the model. Since $\beta\in (0,1)$, we can apply it to the attention computation to reduce the impact of future nodes. Specifically, the modified attention becomes,
\begin{equation}
    Att_N(\bm{h}, U) = softmax(\frac{\bm{q}K}{\sqrt{d}}+log(\bm{\beta_t}))V
\end{equation}
where $\bm{\beta}_t$ is the corresponding gating values $\beta_{u,t}$ of neighbor nodes $u\in U$. When $\beta_{u,t}$ approaches 0, the attention to the corresponding neighbor will become zero, and when $\beta_{u,t}$ approaches 1, the neighbor's attention is computed as normal. Hence, when the gating values are computed correctly, the attention mechanism can select the nodes that reproduce the output like the graph is not incremented with new neighbors. To properly train these gating vectors, we can use a similar supervised signal as in the block-guided loss. For a node $u$ and its corresponding data block $b(u)$, we create a multi-hot vector $l_u$, whose first $b(u)-1$ entries are zeros and last $t-b(u)+1$ are ones, and compute the binary-cross-entropy loss as:
\begin{equation}
    \mathcal{L}_{GBL}(u)=\frac{1}{t}\sum_{j=1}^{t}BinaryCrossEntropy(\beta_{u,j},l_{i,j})
\end{equation}
We term this loss graph block-guided loss(GBL). The loss encourages the gating value to be 1 if the experts are added after the node, and 0 otherwise.
Finally, we need to accommodate the block-guided regularization loss to a more fine-grained version for the interleaving design. Instead of using the target node's corresponding data block as the regularization target, we use each neighbor node's own corresponding data block as the target. And the final loss is,
\begin{equation}
    \mathcal{L}=\mathcal{L}_{cls}+\sum_{i=1}^T\sum_{v\in V}\gamma\mathcal{L}_{BL}(v)+\delta\mathcal{L}_{GBL}(v)
\end{equation}
where $\gamma$ and $\delta$ are hyperparameters. This loss ensures that (a) when the neighbor and the target nodes are from different data blocks, we still want the most relevant expert to be of higher importance; (b) when the correct expert is selected, the expert gets the input that it recognizes from training (low impact from new neighbors, and high impact from old/familiar neighbors).
\subsection{Sparse Dynamic Mixture-of-Experts GNN}
While the proposed DyMoE GNN allows effective knowledge preservation and updates specialized for graph data, it incurs additional computation cost for the dynamically increasing experts. With more data blocks, we can have too many experts whose computational burden overwhelms the performance benefits of the module. Inspired by previous works on Sparse MoE~\citep{shazeer2017outrageously}, we introduce sparsity into the system to improve its efficiency. To that end, we modify Equation~\ref{eq:moe} so that only the experts with the top-k importance score are used to generate predictions. Specifically,
\begin{equation}
\begin{split}
    (\alpha_1...\alpha_t) = Softmax(TopK(s(\bm{x}, \bm{g}_0)...s(\bm{x}, \bm{g}_t))) \\
    \bm{h} = \sum_{i=1}^t\bm{h}_i,\quad \bm{h}_i=\begin{cases}
      \alpha_i f_{\theta_i}(\bm{x}), & \text{if i in TopK} \\
      0, & \text{otherwise}
    \end{cases}
\end{split}
\end{equation}
Because we only use the top-k most essential experts, we do not need to propagate gradients and compute the output of each expert, which significantly reduces the training and inference cost. A complexity analysis can be found in Appendix~\ref{app:sparse}.

Since the last expert and gating are randomly initialized, the model may ignore them because they produce meaningless predictions at the beginning. To mitigate this, we follow Sparse MoE~\citep{shazeer2017outrageously} to tweak the gating values during training randomly so all experts have proper selection chances, and the new experts and gates can gradually learn to correctly predict the new data block. The details about the load balancing can be found in Appendix~\ref{app:sparse}.
\section{Related Work}
\textbf{Incremental Learning} is extensively explored in the deep learning literature, including computer vision~\citep{ewc,lwf,gem} and natural language processing~\citep{nlpcontsurvey, sun2020distill, mi2020continual}. The approaches can be roughly divided into three categories: \textbf{Regularization-based} methods constrain the deviation of the new model from the trained model to retain knowledge~\citep{ewc,zenke2017continual,aljundi2018memory}; \textbf{Experience-Replay} approaches add a small subset of previous data blocks to the current training set as a way to maintain previous knowledge~\citep{gem, er_cont, chaudhry2021using}; \textbf{Architectural} approaches maintain learned knowledge via assigning model parameters to specific data~\citep{aljundi2017expert, ebrahimi2020adversarial, lwf}. Our method falls into the architectural category. Some existing work also considers separate modules for each data block~\cite{aljundi2017expert, rusu2016progressive}, but they focus on the task-incremental scenario, while our method handles both that, and the more challenging class-incremental case. MoE architecture has been applied to solve continual learning problem~\citep{yu2024boosting}, but it does not use data block information to account for structural shift in graph incremental learning, whereas our approach handles this well. 

\textbf{Graph Incremental Learning.} Different from i.i.d. data, graph data suffer from distribution shifts in the incremental learning setting. To overcome this novel challenge, architectural approaches including, PI-GNN~\citep{pi-gnn}, FGN~\citep{wang2022lifelong}, and HPN~\citep{hpn}, use newly initialized model components to learn new knowledge. Experience replay approaches like DyGRAIN~\citep{kim2022dygrain}, ER-GNN~\citep{er}, and Continual GNN~\citep{wang2020streaming} explicitly retrains old nodes selected from graph-related criterion. Regularization approaches such as TWP~\citep{twp}, GraphSail~\citep{graphsail}, GPIL~\citep{tan2022graph}, and SEM~\citep{ricci} identify and minimize a regularization loss to mediate structural shift and correct predictions.  MSCGL~\citep{mscgl} combines architectural search and regularization to preserve learned knowledge. However, because these models treat old models as inseparable units, they ignore different interaction types between data blocks. Meanwhile, our experts are dedicated to individual data blocks, facilitating conditional adaptation to new data. 
\section{Experiments}
\begin{table*}[t]
\setlength{\tabcolsep}{1.5pt}
\caption{AA and AF of class incremental datasets. \textbf{Bold} represents best baseline and \underline{underline} represents runner-up.}
\label{tab:main}
\begin{minipage}[t]{0.99\linewidth}
\resizebox{\linewidth}{!}{%
\begin{tabular}{@{}c|c|ccccccccc|cc@{}}
\toprule
Dataset & Metric & Pretrain & Online & LWF & ER-GNN & SSRM & RCL-CN & PI-GNN & C-GNN & Retrain & DyMoE & DyMoE-L \\ \midrule
& Params. & 4.3M & 4.3M & N/A & 4.3M & 4.3M & 4.7M & N/A & 4.5M & 4.3M & 4.6M & 8.7M \\ \midrule
\multirow{2}{*}{CoraFull} & AA & 17.58{\scriptsize$\pm$1.59 } & 34.37{\scriptsize$\pm$0.69} & 38.61{\scriptsize$\pm$1.28} & 64.57{\scriptsize$\pm$0.60} & 70.71{\scriptsize$\pm$0.59} & 67.37{\scriptsize$\pm$0.62} & 70.92{\scriptsize$\pm$0.60} & 73.98{\scriptsize$\pm$0.61} & 83.07{\scriptsize$\pm$0.79 } & \underline{76.59}{\scriptsize$\pm$0.58} & \textbf{78.16}{\scriptsize$\pm$0.61} \\
 & AF & 0.00{\scriptsize$\pm$0.00 } & -14.64{\scriptsize$\pm$0.79} & -13.39{\scriptsize$\pm$0.98} & -11.92{\scriptsize$\pm$0.57} & \underline{-7.85}{\scriptsize$\pm$0.59} & -11.78{\scriptsize$\pm$0.57} & -9.76{\scriptsize$\pm$0.56} & -9.28{\scriptsize$\pm$0.54} & -0.35{\scriptsize$\pm$0.09 } & -7.92{\scriptsize$\pm$0.54} & \textbf{-7.31}{\scriptsize$\pm$0.57} \\ \midrule
\multirow{2}{*}{Reddit} & AA & 32.19{\scriptsize$\pm$2.93 } & 24.99{\scriptsize$\pm$0.45} & 42.58{\scriptsize$\pm$3.95} & 80.10{\scriptsize$\pm$0.15} & 86.55{\scriptsize$\pm$0.14} & 83.46{\scriptsize$\pm$0.18} & 87.32{\scriptsize$\pm$0.16} & 89.15{\scriptsize$\pm$0.16} & 98.17{\scriptsize$\pm$0.10 } & \underline{92.84}{\scriptsize$\pm$0.13} & \textbf{94.15}{\scriptsize$\pm$0.17} \\
 & AF & 0.00{\scriptsize$\pm$0.00 } & -33.91{\scriptsize$\pm$0.19} & -29.60{\scriptsize$\pm$1.19} & -6.60{\scriptsize$\pm$0.19} & \underline{-2.60}{\scriptsize$\pm$0.20} & -6.36{\scriptsize$\pm$0.19} & -4.36{\scriptsize$\pm$0.23} & -4.07{\scriptsize$\pm$0.21} & 0.14{\scriptsize$\pm$0.04} & \underline{-2.60}{\scriptsize$\pm$0.19} & \textbf{-1.96}{\scriptsize$\pm$0.22} \\ \midrule
\multirow{2}{*}{Arxiv} & AA & 27.91{\scriptsize$\pm$2.47 } & 34.79{\scriptsize$\pm$3.76} & 40.21{\scriptsize$\pm$2.63} & 55.39{\scriptsize$\pm$1.82} & 62.43{\scriptsize$\pm$1.83} & 58.20{\scriptsize$\pm$1.81} & 62.72{\scriptsize$\pm$1.84} & 65.18{\scriptsize$\pm$1.83} & 72.19{\scriptsize$\pm$0.20 } & \underline{68.33}{\scriptsize$\pm$1.80} & \textbf{70.15}{\scriptsize$\pm$1.84} \\
 & AF & 0.00{\scriptsize$\pm$0.00 } & -32.74{\scriptsize$\pm$2.37} & -28.13{\scriptsize$\pm$3.08} & -21.06{\scriptsize$\pm$1.75} & \underline{-9.93}{\scriptsize$\pm$1.73} & -20.81{\scriptsize$\pm$1.73} & -15.37{\scriptsize$\pm$1.72} & -14.44{\scriptsize$\pm$1.75} & 0.47{\scriptsize$\pm$0.18 } & -10.06{\scriptsize$\pm$1.70} & \textbf{-8.82}{\scriptsize$\pm$1.74} \\ \midrule
\multirow{2}{*}{DBLP} & AA & 46.03{\scriptsize$\pm$1.89 } & 47.52{\scriptsize$\pm$3.63} & 50.48{\scriptsize$\pm$3.30} & 54.81{\scriptsize$\pm$3.03} & 56.56{\scriptsize$\pm$3.06} & 55.80{\scriptsize$\pm$3.03} & 56.35{\scriptsize$\pm$3.02} & 57.62{\scriptsize$\pm$3.04} & 65.59{\scriptsize$\pm$1.27 } & \underline{57.75}{\scriptsize$\pm$3.01} & \textbf{58.08}{\scriptsize$\pm$3.05} \\
 & AF & 0.00{\scriptsize$\pm$0.00 } & -17.41{\scriptsize$\pm$2.86} & -14.28{\scriptsize$\pm$2.46} & -8.45{\scriptsize$\pm$0.86} & \underline{-5.43}{\scriptsize$\pm$0.83} & -8.27{\scriptsize$\pm$0.85} & -6.69{\scriptsize$\pm$0.83} & -6.39{\scriptsize$\pm$0.84} & 0.47{\scriptsize$\pm$0.04 } & -5.45{\scriptsize$\pm$0.82} & \textbf{-5.07}{\scriptsize$\pm$0.85} \\ \bottomrule
\end{tabular}}%
\end{minipage}
\vspace{-5pt}
\end{table*}
\begin{table*}[t]
\setlength{\tabcolsep}{2.6pt}
\caption{AA and AF of instance incremental datasets. \textbf{Bold} represents best baseline and \underline{underline} represents runner-up.}
\label{tab:main2}
\begin{minipage}[t]{0.99\linewidth}
\resizebox{\linewidth}{!}{%
\begin{tabular}{@{}c|c|ccccccccc|cc@{}}
\toprule
Dataset & Metric & Pretrain & Online & LWF & ER-GNN & SSRM & RCL-CN & PI-GNN & C-GNN & Retrain & DyMoE & DyMoE-L \\ \midrule
& Params. & 4.3M & 4.3M & N/A & 4.3M & 4.3M & 4.7M & N/A & 4.5M & 4.3M & 4.6M & 8.7M \\ \midrule
\multirow{2}{*}{Paper100M} & AA & 58.61{\scriptsize$\pm$1.98 } & 66.10{\scriptsize$\pm$4.51 } & 74.86{\scriptsize$\pm$2.35 } & 77.25{\scriptsize$\pm$1.32} & 78.92{\scriptsize$\pm$1.31} & 78.09{\scriptsize$\pm$1.29} & 79.13{\scriptsize$\pm$1.31} & 79.94{\scriptsize$\pm$1.31} & 86.15{\scriptsize$\pm$0.49 } & \underline{80.57}{\scriptsize$\pm$1.29} & \textbf{81.24}{\scriptsize$\pm$1.30} \\
 & AF & 0.00{\scriptsize$\pm$0.00 } & -3.97{\scriptsize$\pm$0.43 } & -2.69{\scriptsize$\pm$1.92 } & -4.08{\scriptsize$\pm$0.08} & \underline{-2.03}{\scriptsize$\pm$0.10} & -3.96{\scriptsize$\pm$0.09} & -3.05{\scriptsize$\pm$0.07} & -2.80{\scriptsize$\pm$0.07} & -0.35{\scriptsize$\pm$0.04 } & -2.08{\scriptsize$\pm$0.05} & \textbf{-1.65}{\scriptsize$\pm$0.07} \\ \midrule
\multirow{2}{*}{Elliptic} & AA & 89.91{\scriptsize$\pm$2.41 } & 94.37{\scriptsize$\pm$0.13 } & 94.79{\scriptsize$\pm$0.16 } & 94.37{\scriptsize$\pm$0.05} & 95.11{\scriptsize$\pm$0.02} & 95.17{\scriptsize$\pm$0.05} & \underline{95.67}{\scriptsize$\pm$0.05} & 95.64{\scriptsize$\pm$0.05} & 98.13{\scriptsize$\pm$0.03 } & 95.42{\scriptsize$\pm$0.01} & \textbf{96.12}{\scriptsize$\pm$0.04} \\
 & AF & 0.00{\scriptsize$\pm$0.00 } & -0.98{\scriptsize$\pm$0.88} & -1.96{\scriptsize$\pm$0.14} & -1.03{\scriptsize$\pm$0.19} & \underline{0.10}{\scriptsize$\pm$0.19} & -0.78{\scriptsize$\pm$0.20} & -0.32{\scriptsize$\pm$0.19} & -0.26{\scriptsize$\pm$0.19} & 0.14{\scriptsize$\pm$0.02} & -0.10{\scriptsize$\pm$0.18} & \textbf{0.23}{\scriptsize$\pm$0.21} \\ \midrule
\multirow{2}{*}{Arxiv} & AA & 59.81{\scriptsize$\pm$2.32 } & \underline{69.05}{\scriptsize$\pm$0.39 } & \textbf{70.06}{\scriptsize$\pm$0.64 } & 66.39{\scriptsize$\pm$0.21} & 67.52{\scriptsize$\pm$0.19} & 67.33{\scriptsize$\pm$0.23} & 68.24{\scriptsize$\pm$0.20} & 68.24{\scriptsize$\pm$0.23} & 73.01{\scriptsize$\pm$0.10 } & 68.21{\scriptsize$\pm$0.19} & 68.56{\scriptsize$\pm$0.23} \\
 & AF & 0.00{\scriptsize$\pm$0.00 } & -2.31{\scriptsize$\pm$0.18 } & -1.70{\scriptsize$\pm$0.42 } & -0.23{\scriptsize$\pm$0.37} & \textbf{0.41}{\scriptsize$\pm$0.36} & -0.13{\scriptsize$\pm$0.39} & 0.20{\scriptsize$\pm$0.37} & -0.01{\scriptsize$\pm$0.37} & 0.34{\scriptsize$\pm$0.29 } & -0.01{\scriptsize$\pm$0.35} & \underline{0.24}{\scriptsize$\pm$0.38} \\ \midrule
\multirow{2}{*}{DBLP} & AA & 55.73{\scriptsize$\pm$2.15 } & 63.42{\scriptsize$\pm$1.61 } & 65.15{\scriptsize$\pm$1.76 } & 64.19{\scriptsize$\pm$0.49} & 66.62{\scriptsize$\pm$0.53} & 65.12{\scriptsize$\pm$0.51} & 66.70{\scriptsize$\pm$0.51} & 67.30{\scriptsize$\pm$0.50} & 68.59{\scriptsize$\pm$1.27 } & \underline{67.97}{\scriptsize$\pm$0.49} & \textbf{68.94}{\scriptsize$\pm$0.50} \\
 & AF & 0.00{\scriptsize$\pm$0.00 } & -3.57{\scriptsize$\pm$0.27 } &-2.78{\scriptsize$\pm$0.85 } & -3.74{\scriptsize$\pm$0.74} & -2.51{\scriptsize$\pm$0.76} & -3.73{\scriptsize$\pm$0.76} & -3.07{\scriptsize$\pm$0.74} & -2.94{\scriptsize$\pm$0.77} & 0.29{\scriptsize$\pm$0.04 } & \underline{-2.39}{\scriptsize$\pm$0.73} & \textbf{-2.33}{\scriptsize$\pm$0.76} \\ \bottomrule
\end{tabular}}%
\end{minipage}
\vspace{-5pt}
\end{table*}
We aim to answer the following research questions in the experimental evaluation: \textbf{Q1}: Does the proposed DyMoE framework achieve good empirical performance while maintaining good efficiency? \textbf{Q2}: How does the memory size impact the performance of the model? 
\textbf{Q3}: The framework has several components, how does each component impact its behavior? \textbf{Q4}: Does our training strategy actually encourage dedicated experts? Implementation details and data descriptions can be found in Appedix~\ref{sec:exp}.

\subsection{Quantitative Results}
To answer \textbf{Q1}, we evaluate the model performance with average accuracy (AA) and average forget (AF) on class incremental datasets (CoraFull~\citep{elliptic}, Reddit~\citep{sage}, Arxiv~\citep{ogb}, DBLP-small~\citep{dblp}), and data incremental datasets (Paper100M\citep{ogb}, Elliptic~\citep{elliptic}, Arxiv, DBLP-small). We compared experience-replay baseline (ER-GNN~\citep{er}), architectural baselines (LWF~\citep{lwf}, PI-GNN~\citep{pi-gnn}), and compound baselines (continual-GNN (C-GNN)~\citep{wang2020streaming}, RCL-CN~\citep{rcl-cn}, SSRM~\citep{ssrm}). We also compared with the pretrain baseline, where we only train the model on the first date block and infer all future data blocks; the online baseline, where we directly fine-tune the old model with new data blocks; and the retrain baseline, where we retrain on all data blocks whenever new data blocks arrive. We provide the results of DyMoE module with a similar number of active parameters as baseline methods and a larger version whose individual experts are of same size as the baselines (DyMoE-L). They both have three active experts (k=3). DyMoE represents a fair setting, where our method's and baselines' parameter size and computation time are very close.  We set the same memory node budget for all baselines, the memory budget for each dataset can be found in Appendix~\ref{sec:exp}

We show the experiment results of class incremental setting in Table~\ref{tab:main}, the Params. row shows the active parameters of the baselines, LWF and PI-GNN's parameter sizes can increase indefinitely with the number of datablocks, hence we leave it as "N/A".  From the results, we can see our method significantly improves over most existing baselines for both AA and AF. We reach an average of 3.18\% and 4.92\% relative performance improvement in AA across datasets for DyMoE and DyMoE-L, respectively. The relative improvement is computed by $\frac{AA_{DyMoE}-AA_{B1}}{AA_{B1}}$, where $AA_{B1}$ is the best baseline result. While DyMoE does not reach unanimous superiority on AF, we observe that SSRM, obtaining better AF, has a much lower AA. DyMoE on average improves over SSRM on AA by 6.78\%, whereas the decrease in AF is less than 1\%. Meanwhile, DyMoE outperforms other baselines on AF by large margins. This shows that DyMoE is a more effective trade-off between stability and plasticity. The solid empirical results showed the superiority of the DyMoE design and validated our theory. Comparing DyMoE and DyMoE-L, we see that DyMoE-L achieves better AA and AF, showing that a larger parameter size is more ideal to learn new knowledge without interfering with the old. 
\begin{figure}[t]
    \centering
    \begin{minipage}[t]{0.99\linewidth}
        \includegraphics[width=0.99\linewidth, trim={0.3cm 0.3cm 0.3cm 0.4cm},clip]{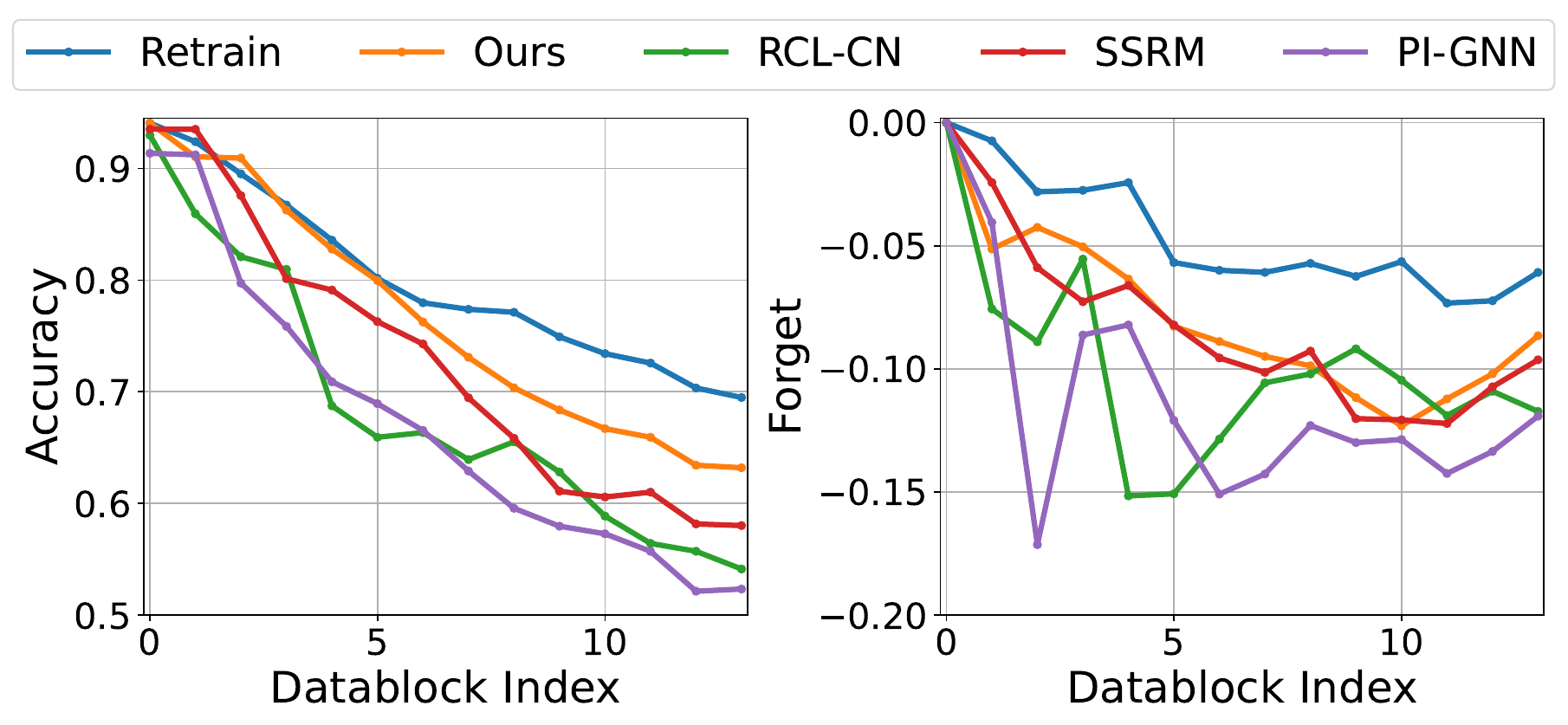}
        \caption{Performance progression on CoraFull dataset over data blocks of our models and baselines.}\label{fig:acc-for}
    \end{minipage}
    \vspace{-5pt}
\end{figure}
\begin{table}
\caption{Training and inference time. (Ave. Seconds/Epoch).}
\label{tab:runtime}
\begin{minipage}[t]{0.99\linewidth}
\resizebox{.98\textwidth}{!}{%
\begin{tabular}{@{}lcccccc@{}}
\toprule
 & \multicolumn{2}{c}{CoraFull} & \multicolumn{2}{c}{Arxiv} & \multicolumn{2}{c}{Reddit} \\ \cmidrule(l){2-7} 
 & \multicolumn{1}{l}{Train} & Inference & \multicolumn{1}{l}{Train} & Inference & \multicolumn{1}{l}{Train} & Inference \\ \midrule
Finetune & 2.41 & 1.49 & 3.59 & 8.96 & 8.79 & 4.57 \\
SSRM & 2.46 & 1.48 & 3.94 & 8.79 & 9.86 & 4.65 \\
PIGNN & 2.95 & 2.06 & 4.59 & 12.59 & 10.57 & 5.91 \\
Retrain & 6.47 & 1.52 & 17.18 & 8.94 & 35.18 & 4.63 \\
DyMoE & 2.47 & 1.55 & 4.17 & 8.86 & 10.28 & 4.75 \\
DyMoE-L & 2.94 & 2.14 & 4.68 & 13.09 & 13.91 & 6.02 \\ \bottomrule
\end{tabular}}%
\end{minipage}
\vspace{-15pt}
\end{table}
For the results of instance incremental learning in Table~\ref{tab:main2}, DyMoE still achieves better results on most targets, including AA on paper100M, Elliptic, and DBLP, while maintaining competitive results on AF. On datasets where DyMoE did not achieve superior results, like Arxiv, we observe that even the most naive baseline (Online) achieved higher performance than other advanced continual learning methods. We suspect that forgetting was not a severe issue in these datasets, and most continual learning methods enforce a mechanism to maintain the old knowledge, which "over-regularizes" the learning process, causing underfitting and degraded performance.

In Table~\ref{tab:runtime}, we show the training and inference time of the baselines and our model. Compared to the retrain baseline (performance upper-bound), both DyMoE and DyMoE-L cost significantly less training time, as the method requires significantly less memory data compared to the retrain method to maintain a competitive performance. DyMoE has very close training and inference time comparing to most effective baselines (SSRM, PIGNN), while achieving better results on most targets. Note that C-GNN is not compared here because it theoretically has the same complexity as Finetune. DyMoE-L, due to a higher parameter size, requires more computation costs during inference, however, DyMoE-L's cost is comparable to that of the architectural approach PIGNN, while maintaining higher performance.
\begin{table}[t]
\setlength{\tabcolsep}{2.6pt}
\caption{Ablation study of class incremental and instance incremental datasets.}
\label{tab:abla2}
\begin{minipage}[t]{0.99\linewidth}
\resizebox{0.99\textwidth}{!}{
\begin{tabular}{@{}lcccccc@{}}
\toprule
 & \multicolumn{2}{c}{Reddit} & \multicolumn{2}{c}{Cora-Full} & \multicolumn{2}{c}{Paper100M}  \\ \cmidrule(l){2-3} \cmidrule(l){4-5} \cmidrule(l){6-7} 
 & AA & AF & AA & AF & AA & AF \\ \midrule
DyMoE & 92.84{\scriptsize$\pm$0.13 } & \textbf{-2.60{\scriptsize$\pm$0.19 }} & 76.59{\scriptsize$\pm$0.58 } & -7.92{\scriptsize$\pm$0.54} & 80.57{\scriptsize$\pm$1.29 } & -2.08{\scriptsize$\pm$0.05 }  \\
DyMoE-$\gamma=0$ & 91.38{\scriptsize$\pm$0.58 } & -3.28{\scriptsize$\pm$0.36 } & 74.39{\scriptsize$\pm$0.58 } & -7.91{\scriptsize$\pm$1.12} & 78.21{\scriptsize$\pm$1.53 } & -3.31{\scriptsize$\pm$1.58 }  \\
DyMoE-$\delta=0$ & 89.94{\scriptsize$\pm$1.28 } & -4.13{\scriptsize$\pm$0.42 } & 72.56{\scriptsize$\pm$0.65 } & -8.47{\scriptsize$\pm$0.58 } & 78.49{\scriptsize$\pm$1.47 } & -3.49{\scriptsize$\pm$1.79 }  \\
DyMoE-$\gamma=0,\delta=0$ & \multicolumn{1}{l}{90.18{\scriptsize$\pm$0.58 }} & \multicolumn{1}{l}{-5.19{\scriptsize$\pm$1.38 }} & \multicolumn{1}{l}{72.17{\scriptsize$\pm$1.14 }} & \multicolumn{1}{l}{-10.62{\scriptsize$\pm$1.20 }} & \multicolumn{1}{l}{76.12{\scriptsize$\pm$0.39 }} & \multicolumn{1}{l}{-4.34{\scriptsize$\pm$1.76 }} \\ \bottomrule
\end{tabular}}
\end{minipage}
\vspace{-10pt}
\end{table}
Furthermore, we plot the AA and AF with respect to the data block sequences in Figure~\ref{fig:acc-for}. We observe that DyMoE can align with the upper-bound retrain method in the first few data blocks in AA, while demonstrating a large margin over other baselines. SSRM achieves better AF, but we can see that in the last few datablocks, DyMoE's forgetting is better, meaning that DyMoE can be more advantageous in maintaining knowledge in a longer data block sequence.

\subsection{Investigation of DyMoE}
\begin{figure}
    \centering
    \begin{minipage}[t]{0.98\linewidth}
        \includegraphics[width=0.99\linewidth]{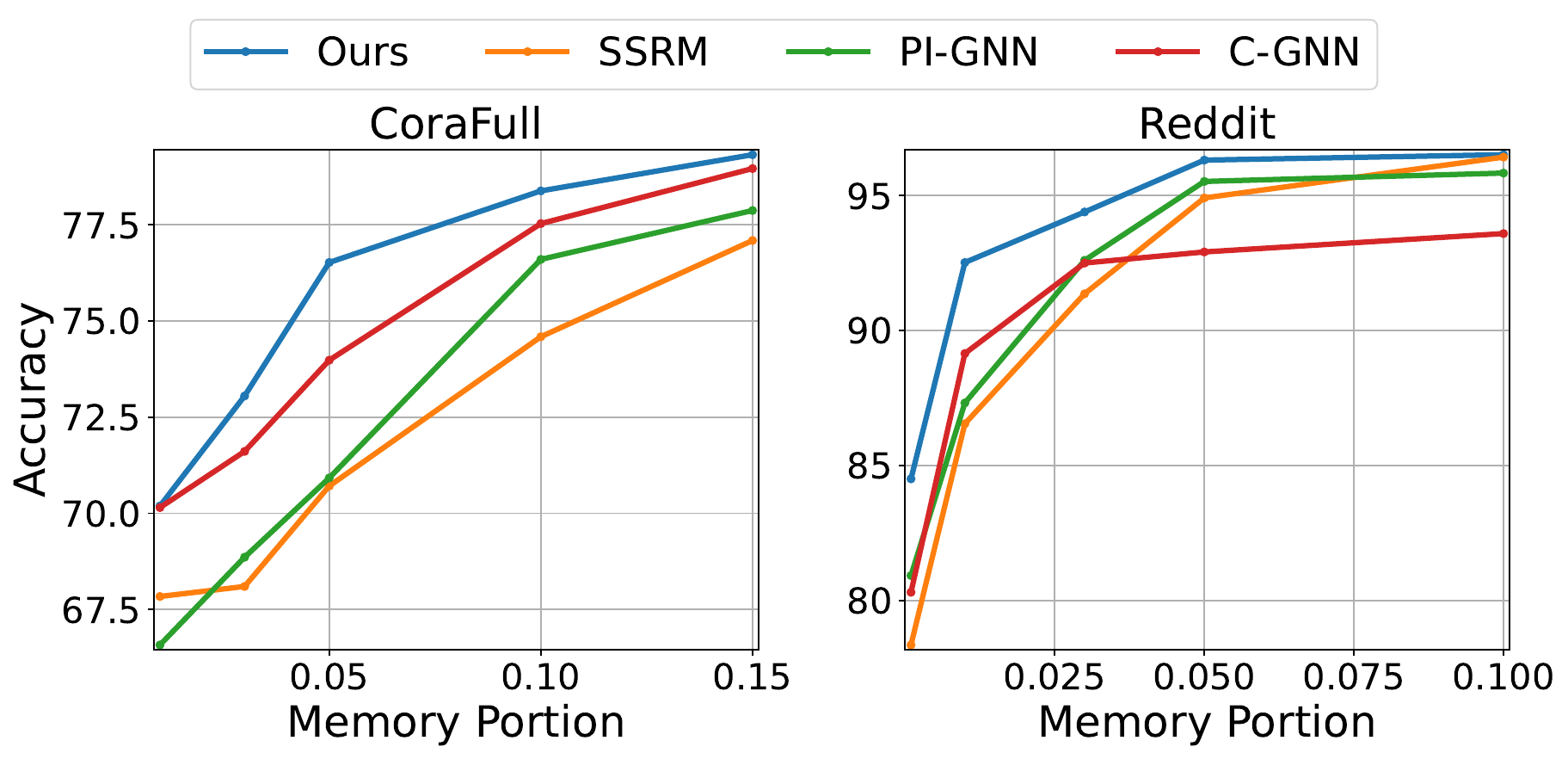}
        \caption{Model accuracy versus memory size for memory-based models.}\label{fig:acc-mem}
    \end{minipage}
    \vspace{-15pt}
\end{figure}

To answer \textbf{Q2}, we compare our method with three other baselines, SSRM, PI-GNN, and C-GNN, that use memory nodes to help retain old knowledge. An ideal incremental learning method should only use a small memory size to obtain desirable performance. We plot the results with different memory portion in Figure~\ref{fig:acc-mem}. From the results, we can see that our approach achieves better performance with the same size of memory, especially when we only have 0.01 memory portion of the training data block on the Reddit dataset. Note that when the memory size approaches infinity, all methods become retrained, and hence we are seeing a converging pattern for the baselines.
\begin{figure}
    \centering
    \includegraphics[width=0.98\linewidth]{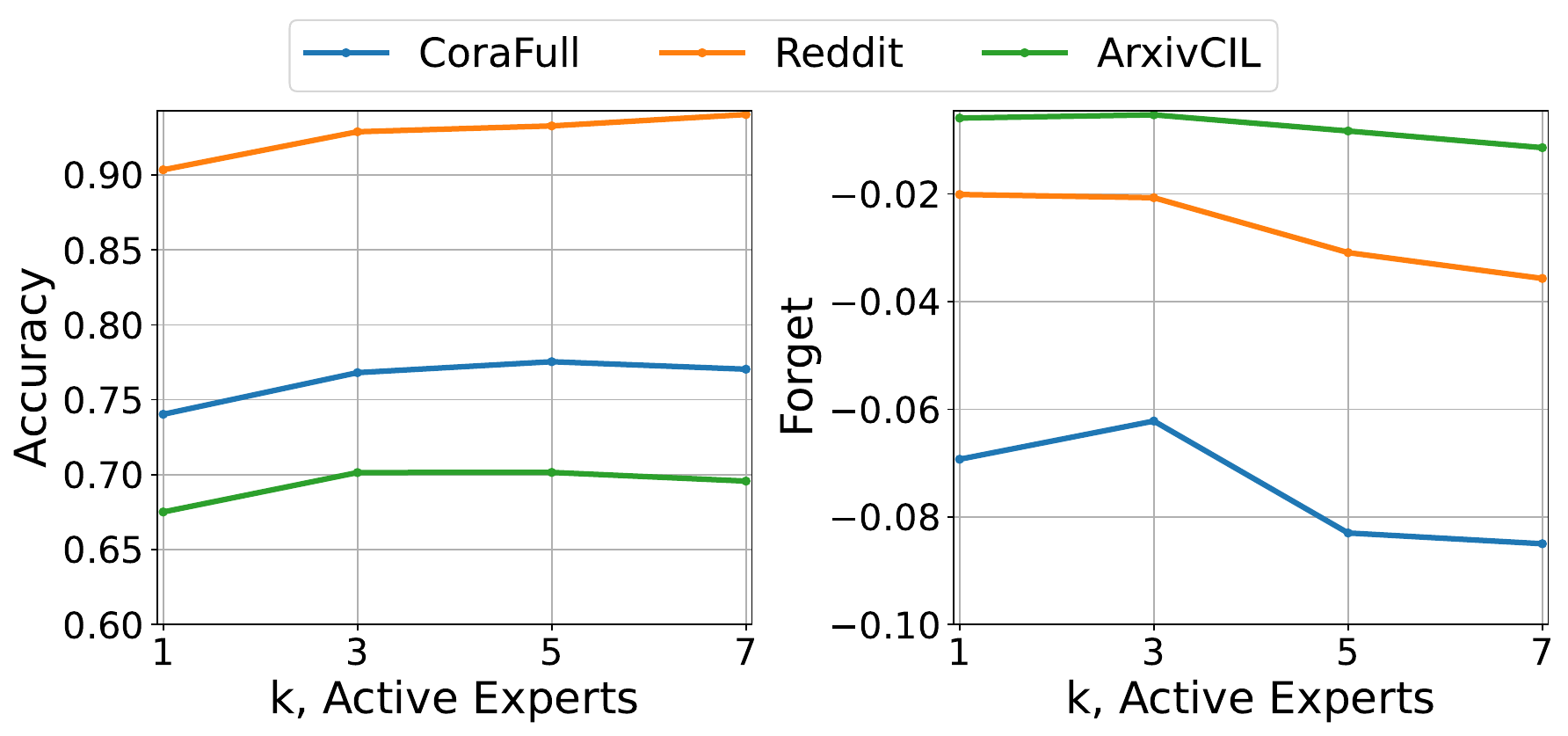}
    \caption{Performance change w.r.t. \# of active experts.}
    \label{fig:kdata}
    \vspace{-15pt}
\end{figure}
To answer \textbf{Q3}, we first investigate the effect of the gating mechanism in DyMoE. To ensure that the nodes are routed to the correct experts and experts only receive nodes that they are familiar with during message-passing, we enforce block-guided loss and graph block-guided loss to provide supervision for the gating mechanism. We study the model performance when these losses are absent from the training process. The results are in Table~\ref{tab:abla2}. We observe that removing either $\gamma$ or $\delta$ leads to performance degradation. When both losses are absent, we usually observe the lowest performance. This shows the necessity of directly using block information as supervision, which is largely ignored in most existing approaches. We also observe that $\delta$ has a stronger impact on the model behavior, as without it the model experiences performance drop, showing the necessity of dedicated experts.

We then investigate how the model reacts to the number of active experts, and show the performance evolution as we increase the number of experts in Table~\ref{fig:kdata}. We can see that the AA increases from $k=1$ to $k=5$, while AF begins to decrease from $k=3$. The divergence between AA and AF after $k=3$ shows that more experts make learning easier, potentially because new experts can borrow more knowledge from the old experts. Meanwhile, it is more difficult to maintain old knowledge. We suspect that this is because extra experts inevitably introduce noises and irrelevant knowledge to old representations, causing forgetting.

\begin{figure}
    \centering
    \includegraphics[width=0.99\linewidth]{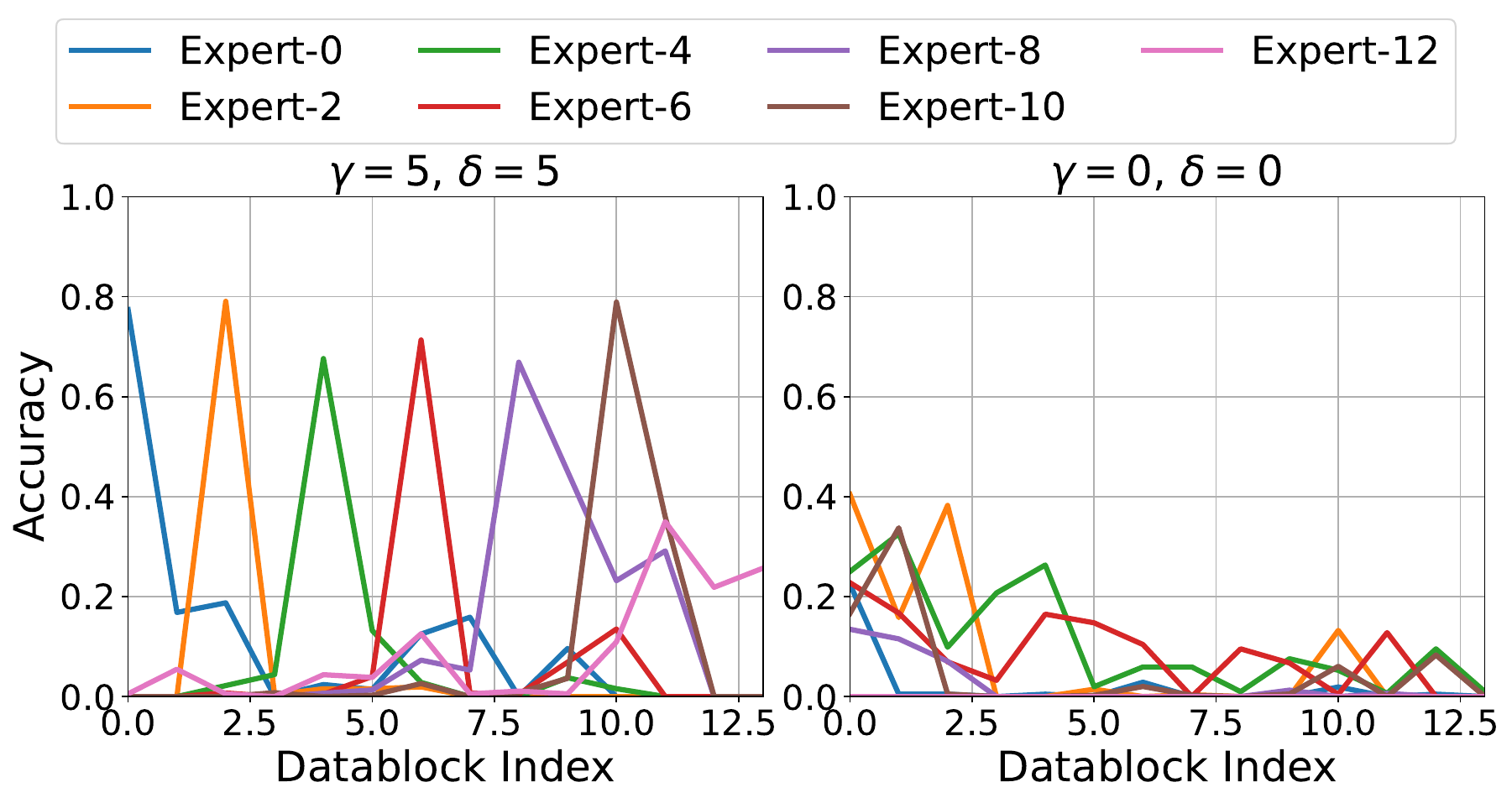}
    \caption{Experts individual performance on data blocks. Each line plot represents an expert.}
    \label{fig:sep-exp}
    \vspace{-15pt}
\end{figure}
\textbf{Q4} validates whether our model and training procedure results in specialized experts as designed. We evaluate the performance of each expert on individual data blocks. In Figure~\ref{fig:sep-exp}, we can easily observe that when trained with block-guided loss, the experts are specialized, and they achieve high prediction accuracy for their corresponding data blocks. When the block-guided loss is absent, the experts fail to specialize, further demonstrating the necessity of using block-information to guide training the MoE for continual learning.
\section{Conclusion, Limitations, and Future Work}
In this paper, we identified the drawbacks of existing graph incremental learning models and proposed the DyMoE module with a sparse version to model different interaction types between data blocks effectively and efficiently. However, we also acknowledge that our model may have trouble locating the correct experts when there are too many data blocks, resulting in compromised performance. While this can be solved by periodic retraining, we plan to extend our work to handle extremely long data sequences (over 1000 data blocks) in future work.
\bibliographystyle{ACM-Reference-Format}
\bibliography{iclr2025_conference}

\newpage

\appendix

\section{Proof of Theorem 1}\label{sec:app_proof}
We restate Theorem~\ref{thm:main_thm} for completeness,
\mainthm*

It is easy to see that DyMoE is at least as powerful as PI since we can parameterize all gating vectors with the same value; hence, the weights of all experts are the same, which makes the final output essentially a summation of each expert's output. In this case, DyMoE degenerates to PI. We then prove that under the Gaussian Mixture assumption of the data blocks, DyMoE achieves lower loss and, hence, is strictly more powerful than PI.
\begin{proof}
    Consider the case with two data blocks generated from Gaussian Distributions, $X_1= \mathcal{N}(\bm{\mu}_1,\sigma^2 I)$, and $X_2= \mathcal{N}(\bm{\mu}_2,\sigma^2 I)$. For simplicity, we assume the same variance across coordinates and the probability of data from each distribution is the same. Let the distance between two distributions be $B$. The labels of data are generated depending on their distance from the mean of their coorresponding distribution, specifically, if $\bm{x}\sim X_1$,
    \begin{equation}
    y= 
    \begin{cases}
        0,& \text{if } ||\bm{x}-\bm{\mu}_1||\leq d\\
        1,              & \text{otherwise}
    \end{cases}
    \end{equation}
    and if $x\sim X_1$,
    \begin{equation}
    y= 
    \begin{cases}
        2,& \text{if } ||\bm{x}-\bm{\mu}_2||\leq d\\
        3,              & \text{otherwise}
    \end{cases}
    \end{equation}
    where $2d\leq B$ is a threshold distance to determine the data labels. This is a practical assumption for a mixture of two Gaussian distributions.

    We then consider the procedure of Parameter Isolation (PI) and our proposed method. PI first trains a model $f_1(x)$ on $X_1$ and then trains a model $f_2(x)$ on $X_2$, both $f_1$ and $f_2$ are in $R^4$ for the four target classes. Hence, when making predictions, we have the logits to be:
    \begin{equation}
        y=\text{softmax}(f_1(x)+f_2(x))
    \end{equation}
    Since our approach can initialize a network with the same architecture, we can have the same network and parameters as the ones in PI, and the predictions from our model are:
    \begin{equation}
    \begin{split}
        y=\text{softmax}(\alpha_1f_1(\bm{x})+\alpha_2f_2(\bm{x})),\\
        \alpha_i=\frac{exp(-\frac{||\bm{x}-\bm{g}_i||^2}{2\sigma^2})}{exp(-\frac{||\bm{x}-\bm{g}_1||^2}{2\sigma^2})+exp(-\frac{||\bm{x}-\bm{g}_2||^2}{2\sigma^2})}
    \end{split}
    \end{equation}
    Here, we use the negative of the distance normalized by the variance as the similarity measure between the input and the gating vectors. This is a valid and tractable choice as variance can be estimated by batch normalization. Note that in the first data block, we directly set the gating vector to the empirical mean of the input, $\bm{g}_1=\bar{\bm{x}}_1\approx\bm{\mu}_1$. In the second data block, the block-guided loss solves the problem
    \begin{equation}
        \min_{\bm{g}_2} \frac{1}{N}\sum_{i=1}^N s(\bm{x}_i, \bm{g}_2)
    \end{equation}
    which is minimized by $\bm{g}_2=\bm{\mu}_2$. We can then rewrite the prediction of the model:
    \begin{equation}
    \begin{split}
        y=\text{softmax}(\alpha_1f_1(\bm{x})+\alpha_2f_2(\bm{x})),\\
        \alpha_i=\frac{exp(-\frac{||\bm{x}-\bm{\mu}_i||^2}{2\sigma^2})}{exp(-\frac{||\bm{x}-\bm{\mu}_1||^2}{2\sigma^2})+exp(-\frac{||\bm{x}-\bm{\mu}_2||^2}{2\sigma^2})}
    \end{split}
    \end{equation}
    To show that our model achieves lower loss on this task, we only need to consider the expected loss on the $D_1$ as the two distributions are symmetric. We divide the problem into two cases $||\bm{x}-\bm{\mu}_1||\leq d$, when the input is close to the distribution mean, and $||\bm{x}-\bm{\mu}_1||> d$ when the input is farther away.

    For $||\bm{x}-\bm{\mu}_1||\leq d$, the correct label is $0$. We consider the cross-entropy loss of PI,
    \begin{equation}
    \begin{split}
        \mathcal{L}_{PI}&=-log(\frac{f_1(\bm{x})_0}{f_1(\bm{x})_0+f_1(\bm{x})_1+f_2(\bm{x})_2+f_2(\bm{x})_3})\\
        &=-log(\frac{\alpha_1f_1(\bm{x})_0}{\alpha_1f_1(\bm{x})_0+\alpha_1f_1(\bm{x})_1+\alpha_1f_2(\bm{x})_2+\alpha_1f_2(\bm{x})_3})
    \end{split}
    \end{equation}
    and the cross-entropy loss of our method
    \begin{equation}
        \mathcal{L}_{Dy}=-log(\frac{\alpha_1f_1(\bm{x})_0}{\alpha_1f_1(\bm{x})_0+\alpha_1f_1(\bm{x})_1+\alpha_2f_2(\bm{x})_2+\alpha_2f_2(\bm{x})_3})
    \end{equation}
    since $||\bm{x}-\bm{\mu}_1||\leq d\leq B-d\leq ||\bm{x}-\bm{\mu}_2||$, meaning that $\alpha_1 \geq \alpha_2$, and we have $\mathcal{L}_{Dy}\leq \mathcal{L}_{PI}$. Since cross-entropy is monotonic, we can obtain the minimum of $\mathcal{L}_{PI}-\mathcal{L}_{Dy}$ at $||\bm{x}-\bm{\mu}_1||=d$ and $||\bm{x}-\bm{\mu}_2||=B-d$. Let the minimum be $\Delta\mathcal{L}_{close}$. Note $\Delta\mathcal{L}_{close}$ increases as $d$ increases and $\sigma$ decreases.

    For $||\bm{x}-\bm{\mu}_1||> d$, the correct label is $1$. Let $M$ be the maximum absolute value that the neural network $f_2$ output for a logit, that is $|f_2(x)_y|\leq M$. Then, the maximum possible loss is $-f_2(x)_1+log(C\cdot exp(M))=log(C) + 2M$, where $C=4$, the number of classes. Since logits of PI and our method is bounded by the same $M$, we have the maximum possible loss difference to be 
    \begin{equation}
       |\Delta \mathcal{L}_{far}| = 4M+2log C
    \end{equation}

    We now developed a lower bound for the loss difference when $\bm{x}$ is close to $\bm{\mu}_1$ and an upper bound for the loss difference when x is far from $\bm{\mu}_2$, we then compute the probability of each case using Gaussian tail bound.
    \begin{equation}
        P[\bm{x}-\bm{\mu}_1>d]\leq exp(-\frac{d^2}{2\sigma^2}), P[\bm{x}-\bm{\mu_1}\leq d]\geq 1 - exp(-\frac{d^2}{2\sigma^2})
    \end{equation}
    Then the upper bound of the difference in expected loss when $x$ is far is:
    \begin{equation}
        |\Delta E_{far}|\leq (4M+2logC)\cdot exp(-\frac{d^2}{2\sigma^2})
    \end{equation}
    The lower bound of the expected loss when $x$ is close is:
    \begin{equation}
        \Delta E_{close}\geq \Delta\mathcal{L}_{close}\cdot (1-exp(-\frac{d^2}{2\sigma^2}))
    \end{equation}
    Taking the ratio:
    \begin{equation}
        \frac{|\Delta E_{far}|}{\Delta E_{close}}\leq\frac{(4M+2logC)\cdot exp(-\frac{d^2}{2\sigma^2})}{\Delta\mathcal{L}_{close}\cdot (1-exp(-\frac{d^2}{2\sigma^2}))}
    \end{equation}
    As $\frac{d}{\sigma}$ increases, the ratio approaches zero, hence we have the overall expected loss difference,
    \begin{equation}
        \Delta E=\Delta E_{close}+|\Delta E_{far}|\geq\Delta E_{close}-|\Delta E_{far}|\geq0
    \end{equation}
    making the overall loss difference positive, and our approach leads to lower loss in this case.
\end{proof}

\section{Memory Set Construction}\label{app:mem}
DyMoE utilizes memory sets to train gating vectors for correct data routing. To construct the memory, we first set a memory budget $0<p<1$, representing the portion of the full data block $X$ that will be kept as memory set $M$, that is
\begin{equation}
    |M|=p|X|
\end{equation}
Usually, $p$ is a small value ($p<0.05$) to ensure the efficiency. Inspired by ER-GNN~\cite{er}, we use a sample's representativeness to select memory nodes. Let $X_c$ be all samples in $X$ that has class $c$, we collect their learned representation before the final logit prediction layer, $F_c={f(x)|x\in X_c}$. We then compute the representative vector $\bm{x}_c$ as
\begin{equation}
    \bm{x}_c=\frac{1}{|F|}\sum_{\bm{c}\in F}\bm{c}
\end{equation}
Then representativeness of sample $\bm{x}$ is determined by the norm distance between the sample representation and the representative vector, $s=-||f(\bm{x})-\bm{x}_c||$. We sort this value in $X_c$, and pick the largest $k_c$ samples to add to the memory set. $k_c$ is determined by the class distribution.
\begin{equation}
    k_c=\frac{|X_c|}{|X|}
\end{equation}
A benefit of this construction strategy is that the resulting memory set $\bigcup_i M^{(i)}$ reflects the actual class distribution but with much less data. Hence, we can use it to balance the data during training. The training is divided into two stages, in the first stage, the model is trained with $\bigcup_{i=1}^{t-1}M^{(i)}\cup X^{(t)}$, so that the last expert can learn fine-grained information from the full new data block. Then, we collect memory for $X^{(t)}$ as described above and train the model with $\bigcup_{i=1}^{t}M^{(i)}$, this teaches the gating mechanism the correct class distribution of the current graph.
\section{Details about Sparse DyMoE}\label{app:sparse}
Since DyMoE in each individual timestamp can be interpreted as a conventional static sparse MoE model, we can apply the same strategy~\cite{shazeer2017outrageously} to ensure experts get a good chance of being selected, including the randomly and newly initialized ones. Specifically, 
\begin{equation}
\begin{split}
     (\alpha_1,...,\alpha_t)=Softmax(KeepTopK(H(\bm{x},\bm{g}_1,\bm{q}_1),...,H(\bm{x},\bm{g}_t,\bm{q}_t))\\
     H(x,g_i,q_i)= s(\bm{x},\bm{g}_i)+StandardNormal()\cdot Softplus(s(\bm{x}, \bm{q}_i))\\
     KeepTopK(v_1,...,v_t)=\begin{cases}
      v_i, & \text{if i in TopK} \\
      -\inf, & \text{otherwise}
    \end{cases}
\end{split}
\end{equation}
We add a noise term whose magnitude is determined by another learnable noise vector.

Modern MoE designs also employ load balancing design to ensure each experts get similar number of samples. However, we observe that such a strategy is not bringing performance boost to our methods, potentially because the supervised data-block signal already handles the load balancing issue.
\section{Experiment Details}\label{sec:exp}
\subsection{Implementation Details}
\begin{table*}[t]
\centering
\caption{Hyperparameters for class incremental learning.}\label{tab:hyper1}
\begin{tabular}{@{}lcccc@{}}
\toprule
 & Arxiv-CIL & DBLP-CIL & CoraFull & Reddit \\ \midrule
Learning Rate & \multicolumn{4}{c}{0.0001} \\
Weight Decay & \multicolumn{4}{c}{\{0.01, \textbf{0.001}, 0.0001\}} \\
Embedding Dimension & \multicolumn{4}{c}{128} \\
\# Epochs & \multicolumn{4}{c}{40} \\
\# Balancing Epochs & \multicolumn{4}{c}{10} \\
$\gamma$ & \{0.01, 0.1, \textbf{1}, 5\} & \{0.01, \textbf{0.1}, 1, 5\} & \{0.01, 0.1, 1, \textbf{5}\} & \{0.01, 0.1, \textbf{1}, 5\} \\
$\delta$ & \multicolumn{4}{c}{5} \\
$p$ & 0.01 & 0.01 & 0.05 & 0.01\\
batch size & \multicolumn{4}{c}{128} \\ \bottomrule
\end{tabular}
\end{table*}

\begin{table*}[t]
\centering
\caption{Hyperparameters for instance incremental learning.} \label{tab:hyper2}
\begin{tabular}{@{}lcccc@{}}
\toprule
 & Arxiv-IIL & DBLP-IIL & \multicolumn{1}{c}{Paper100M} & \multicolumn{1}{c}{Elliptic} \\ \midrule
Learning Rate & \multicolumn{4}{c}{0.0001} \\
Weight Decay &  \multicolumn{4}{c}{0.001} \\
Embedding Dimension & \multicolumn{4}{c}{128} \\
\# Epochs & \multicolumn{4}{c}{40} \\
\# Balancing Epochs & \multicolumn{4}{c}{5} \\
$\gamma$ & \{0.01, \textbf{0.1}, 1, 5\} & \{\textbf{0.01}, 0.1, 1, 5\} & \multicolumn{1}{c}{\{0.01, \textbf{0.1}, 1, 5\}} & \multicolumn{1}{c}{\{0.01, \textbf{0.1}, 1, 5\}} \\
$\delta$ & \multicolumn{4}{c}{5} \\
$p$ & \multicolumn{4}{c}{0.01}\\
batch size & \multicolumn{4}{c}{128} \\ \bottomrule
\end{tabular}
\end{table*}
The repository for implementation can be found at the following \href{https://github.com/amazon-science/dymoe-graph-incremental-learning}{\color{blue}https://github.com/amazon-science/dymoe-graph-incremental-learning}. 
The model is implemented in PyTorch and DGL, and all experiments are conducted on 1 Nvidia A100 80GB GPU. We repeat the experiment 5 times using different random seeds and report the mean and standard deviation. We uniformly use a fan-out of 10 to extract subgraphs from each target node. The hyperparameters used during training are shown in Table~\ref{tab:hyper1} and Table~\ref{tab:hyper2}, where the curly bracket represents the hyperparameters for searching, and the hyperparameters selected are marked in bold. Memory size is the per data block memory size, and it is a special hyperparameter in the continual learning setting because as it increases, all methods converge to the retrain method, which is usually the upper bound of all continual learning methods. We set a uniform ratio $p$ of the training dataset for all methods use memory set.

\subsection{Dataset Details}
\begin{figure*}
    \centering
    \includegraphics[width=\linewidth]{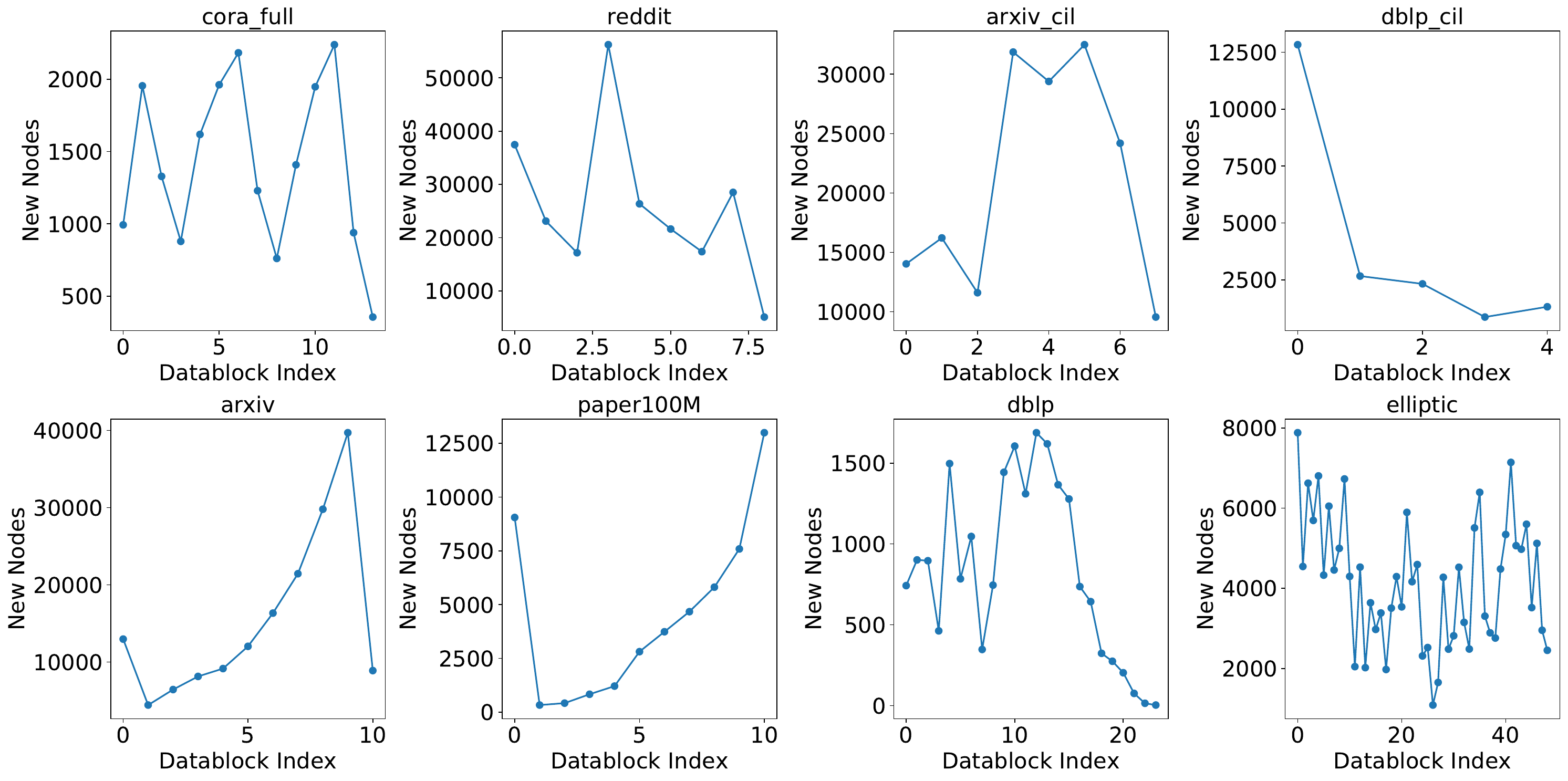}
    \caption{Number of new nodes per data block.}
    \label{fig:new-node}
\end{figure*}

\begin{figure*}
    \centering
    \includegraphics[width=\linewidth]{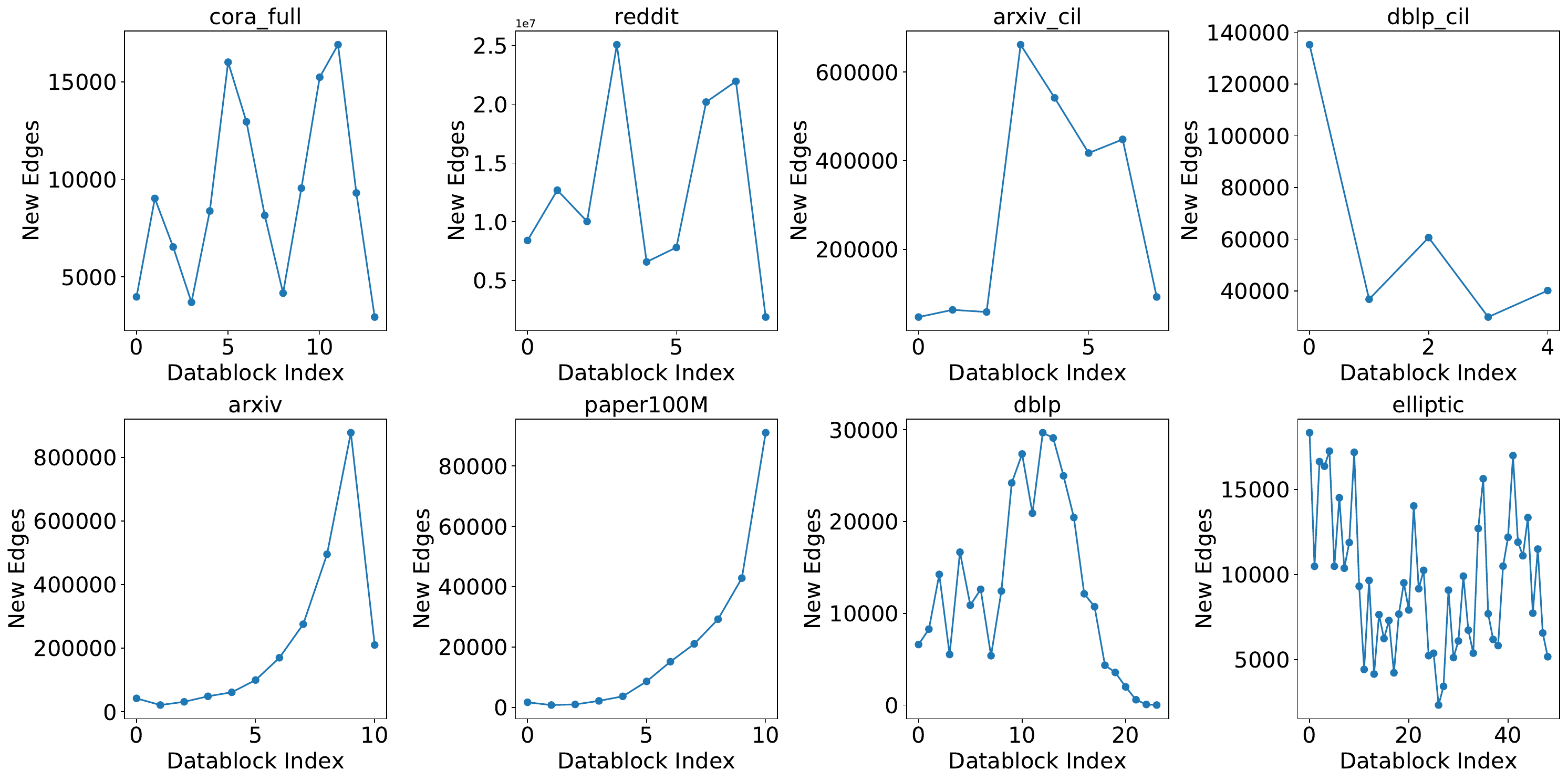}
    \caption{Number of new edges per data block.}
    \label{fig:new-edge}
\end{figure*}

\begin{table*}[t]
\caption{Dataset statistics.}
\label{tab:datastat}
\begin{tabular}{@{}lccccc@{}}
\toprule
 & \#. Nodes & \#. Edges & \#. Classes & \#. Data blocks & \#. Classes per block \\ \midrule
CoraFull & 19793 & 126842 & 70 & 14 & 5 \\
Arxiv-CIL & 169343 & 2332486 & 40 & 8 & 5 \\
Reddit & 232965 & 114615892 & 41 & 9 & 5 \\
DBLP-small-CIL & 20000 & 302862 & 9 & 5 & 2 \\
Paper100M-small & 49459 & 217420 & 12 & 11 & NA \\
Arxiv-IIL & 169343 & 2332486 & 40 & 11 & NA \\
DBLP-small-IIL & 20000 & 302826 & 9 & 24 & NA \\
Elliptic & 203769 & 468710 & 2 & 49 & NA \\ \bottomrule
\end{tabular}
\end{table*}
The dataset statistics are shown in Table~\ref{tab:datastat}. We collect data from academic graphs (Arxiv, DBLP, Paper100M, CoraFull), social networks (Reddit), and BlockChain networks (Elliptic) to show that our model handles a wide range of datasets. We describe the construction of each dataset as follows and includes the number of new nodes and edges in Figure~\ref{fig:new-node} and \ref{fig:new-edge}. We use the provided train/valid/test split from the dataset source. If the source does not have established split, we use 60/20/20 train valid test split.

\textbf{ArXiv}: Arxiv academic citation network from Open Graph Benchmark (OGB)~\citep{ogb} contains arxiv articles and the citation information between articles. For instance incremental learning setting, we use the first 25 timestamps in the original arxiv dataset as the first data block, as they contain significantly less data. We then split the rest of the data by year, and data in each forms a data block. For class incremental learning setting, we split the data into 8 blocks each contains 5 classes.

\textbf{DBLP}: DBLP is an academic network from the DBLP website containing computer science academic paper, with citation information~\citep{dblp}. We follow \citet{pi-gnn} to sample 20000 nodes with 9 classes and 75706 edges from DBLP full data, we split it into data blocks according to the timestamps. For the class incremental setting, we split the 9 classes into 5 data block each containing 2 classes, except for the last one with only 1 class.

\textbf{Paper100M}: Paper100M is a citation network extracted from Microsoft Academic Graph by OGB~\citep{ogb}. We follow \citet{pi-gnn} to sample 12 classes from the year 2009 to the year 2019 from Paper100M full data and we split it into tasks according to the timestamps.

\textbf{CoraFull}: CoraFull is a co-citation academic network, where nodes are papers, and the two nodes are connected if they are co-cited by other papers~\citep{corafull}. We use the provided CoraFull data from DGL, and split its 70 classes into 14 5-classes data blocks for class incremental learning.

\textbf{Reddit}: The Reddit dataset contains Reddit posts as nodes, and two nodes are connected by edges if they are posted by the same user~\citep{sage}. We use the provided Reddit data from DGL, and split its 40 classes into 8 5-classes data blocks for class incremental learning.

\textbf{Elliptic}: The Elliptic dataset is a bitcoin transaction network, where each node represents a transaction, and each edge denotes money flow~\citep{elliptic}. Its nodes have timestamps evenly spaced with an interval about two weeks. We use the original timestamp from the dataset for instance-incremental learning.


\end{document}